\documentclass[conference,usletter]{IEEEtran}
\IEEEoverridecommandlockouts
\usepackage[utf8]{inputenc}
\usepackage[T1]{fontenc}
\usepackage{cite}
\usepackage{amsmath,amssymb,amsfonts}
\usepackage{accents}
\usepackage{bm}
\usepackage{algorithmic}
\usepackage{graphicx, tabularx, cellspace}
\usepackage{xcolor}
\usepackage{multirow}
\usepackage{url}
\usepackage{soul,soulpos}
\usepackage{placeins}

\def\BibTeX{{\rm B\kern-.05em{\sc i\kern-.025em b}\kern-.08em
 T\kern-.1667em\lower.7ex\hbox{E}\kern-.125emX}}
 
\newcommand{\ner}[2]{\underaccent{\text{#1}}{\underline{\text{#2}}}}

\makeatletter
\newcommand*\titleheader[1]{\gdef\@titleheader{#1}}
\AtBeginDocument{%
  \let\st@red@title\@title
  \def\@title{%
    \bgroup\vskip-1.7em\normalfont\large\@titleheader\par\egroup
    \vskip.4em\st@red@title}
}
\makeatother

\title{Cyberthreat Detection from Twitter using\\
Deep Neural Networks}
\titleheader{\copyright~IEEE 2019\\Accepted for the 2019 International Joint Conference on Neural Networks}

\begin{document}
%\thanks{This work was supported by the European Commission through the H2020 DiSIEM ref. G.A. 700692, and by the FCT through the Research Unit LASIGE, ref. UID/CEC/00408/2013.}
%}
\author{
\IEEEauthorblockN{
Nuno Dionísio,
Fernando Alves, 
Pedro M. Ferreira and
Alysson Bessani}
\IEEEauthorblockA{LASIGE, Faculdade de Ciências, Universidade de Lisboa\\
Lisboa 1749-016, Portugal\\
Email: \{ndionisio, falves\}@lasige.di.fc.ul.pt,
\{pmf, anbessani\}@ciencias.ulisboa.pt}}

%%======================================================================================
%% Copyright do PDF
%\IEEEpubid{
%\begin{minipage}
%{\textwidth}\ \\[12pt]
%\copyright 2019 IEEE\\ International Joint Conference on Neural Networks
%\end{minipage}
%} 
%%======================================================================================
%%======================================================================================
% Atual
%\IEEEpubid{
%\begin{minipage}%
%{\textwidth}\ \\[12pt]
%\copyright 2019 IEEE\\ Accepted for 2019 International Joint Conference on Neural Networks
%\end{minipage}
%} 
%%======================================================================================
\maketitle

\begin{abstract}
To be prepared against cyberattacks, most organizations resort to security information and event management systems to monitor their infrastructures.
These systems depend on the timeliness and relevance of the latest updates, patches and threats provided by cyberthreat intelligence feeds.
Open source intelligence platforms, namely social media networks such as Twitter, are capable of aggregating a vast amount of cybersecurity-related sources.
To process such information streams, we require scalable and efficient tools capable of identifying and summarizing relevant information for specified assets.
This paper presents the processing pipeline of a novel tool that uses deep neural networks to process cybersecurity information received from Twitter. 
A convolutional neural network identifies tweets containing security-related information relevant to assets in an IT infrastructure.
Then, a bidirectional long short-term memory network extracts named entities from these tweets to form a security alert or to fill an indicator of compromise.
The proposed pipeline achieves an average 94\% true positive rate and 91\% true negative rate for the classification task and an average F1-score of 92\% for the named entity recognition task, across three case study infrastructures.
\end{abstract}


\section{Introduction}
Cybersecurity is becoming an ever-increasing concern for most organizations, with studies estimating the cost of cybercrime to be up to 0.8 percent of the global GDP~\cite{CSIS}.
A Security Operations Center (SOC) requires timely and relevant threat intelligence to accurately and properly monitor, maintain, and secure an IT infrastructure.
This leads security analysts to strive for threat awareness by collecting and reading various information feeds.
However, if done manually, this results in a tedious and extensive task that may result in little knowledge being obtained given the large amounts of irrelevant information an analyst may have to go through.
Although there is the option of subscribing a paid service to receive curated feeds, research has shown that Open Source Intelligence (OSINT) provides useful information to create Indicators of Compromise (IoC)~\cite{paper_SONAR,paper_IOC_game,paper_Sabottke}.

In this paper, we present a threat intelligence tool that employs deep neural network architectures to process a data stream, identifying relevant security-related information and extracting relevant entities, thus synthesizing valuable knowledge that can be used by a SOC. 

Although other text-based sources could leverage our approach, we opted to focus on Twitter due to its ability to act as a natural aggregator of multiple sources~\cite{Obe17}.
This social media platform offers a large and diverse pool of users, high accessibility, timeliness, thus producing a large volume of data.
These properties remain true regarding the cybersecurity field~\cite{paper_Sabottke}.
From security researchers, companies and enthusiasts to hacker groups, there is a rich and timely flow of security-related data.

The tool proposed in this paper collects tweets from a selected subset of accounts using the Twitter streaming API, and then, by using keyword-based filtering, it discards tweets unrelated to the monitored infrastructure assets.
To classify and extract information from tweets we use a sequence of two deep neural networks.
The first is a binary classifier based on a Convolutional Neural Network (CNN) architecture used for Natural Language Processing (NLP)~\cite{paper_kim}.
It receives tweets that may be referencing an asset from the monitored infrastructure and labels them as either relevant when the tweets contain security-related information, or irrelevant otherwise.
Relevant tweets are processed for information extraction by a Named Entity Recognition (NER) model, implemented as a Bidirectional Long Short-Term Memory (BiLSTM) neural network~\cite{paper_BiLSTMCRF}.
This network labels each word in a tweet with one of six entities used to locate relevant information.

%\subsection{Why Deep Learning?}
% Movi esta parte para um paragrafo acima
%There are two main reasons why we are employing deep learning techniques at the core of our tool: one motivated by the desire to have an appropriate architecture with task-related functional layers, the other related to the NLP nature of the task.
We are seeking a complete end-to-end architecture with no requirement for feature engineering and extra components in the processing pipeline.
Furthermore, we have chosen to pursue the application of deep learning techniques because of its advantages in the NLP domain~\cite{Cun15}.
% Contributions
%\subsection{Paper contributions}
Thus, we propose an end-to-end threat intelligence tool which relies on neural networks with no feature engineering. 
The pipeline is capable of receiving tweets relevant to an infrastructure, selecting those which appear to contain relevant information regarding an asset's security and extracting valuable entities which can be used to issue a security alert or to improve an IoC.

To evaluate our approach we collected tweets corresponding to three IT infrastructures defined by three private organizations: a nation-wide power utility, a large company's cybersecurity consultancy department, and a worldwide travel services provider.
%We establish a methodology through which we compare several variations to the deep learning architectures in order to select a model which presents the best performance according to a defined evaluation metric.
We establish a methodology through which, according to a defined evaluation metric, we compare several variations to the deep learning architectures to select a model which provides the best performance.
Furthermore, we compare the proposed models to other well-known classifiers and provide a detailed analysis of the results obtained.

The evaluation shows that our approach is capable of finding, on average, more than 92\% of the relevant tweets, and matching cybersecurity-relevant labels to named entities within these tweets with an average F1-score above 90\%.
Based on the best models obtained in our experiments, we retrieved tweets where the NER models were capable of extracting relevant entities and performed a brief analysis which demonstrates the timeliness of Twitter as a valuable source for relevant cyberthreat awareness.

\section{Related Work}
In this section, we review previous work that used Twitter or other OSINT sources for cybersecurity, and employed machine learning for cyberthreat detection.
Additionally, we review previous research on deep learning techniques to process tweets and perform NER or other NLP tasks.

\subsection{Machine learning and OSINT for cyberthreat detection}
Previous work has explored the possibility of using Twitter as an information source for cyberthreat awareness.
Le Sceller et al.~\cite{paper_SONAR} proposed SONAR, an automatic keyword-centric self-learned framework that can detect, geolocate and categorize cybersecurity events in near real-time, based on a Twitter stream.
The authors describe the framework in three phases.
The first is similar to ours, as the framework queries Twitter with a list of keywords to retrieve a continuous stream of tweets.
However, instead of querying Twitter as a whole, we focus on a pre-defined set of accounts that are more likely to tweet security-related content, thus avoiding possible performance decrease due to noisy content.
Moreover, the keywords we use are not cyberthreat related (e.g., attacks and vulnerabilities).
Instead, we query for a list of keywords that define the assets in the monitored infrastructure.
%Our method of tweet extraction differ from this work because it is focused on using keywords which describe assets of a given infrastructure instead of using cyberthreat specific keywords.
%Moreover, we focus on a pre-defined set of accounts that are more likely to tweet security-related content, thus avoiding possible performance decrease due to noisy content.
The second phase of SONAR is about event detection. 
A clustering technique is used to aggregate similar tweets that may report the same cybersecurity event.
%These events are geolocated, classified and displayed on a user interface. 
Given that the system relies on keywords being up-to-date, the third phase aims to find new keywords based on their co-occurrence with other previously defined tweets. 
These two phases differ from ours in scope, since our aim is not to build a general information security news feed but rather a tool to gather the most recent information related to the security of a pre-defined list of assets.
%In general, SONAR strays from our scope, since our aim is not to build a general information security news feed but rather a tool to actively gather information related to the potential security risks to an infrastructure defined by a list of assets.

Sabottke et al.~\cite{paper_Sabottke} proposed a Twitter-based exploit detector using an SVM classifier.
The detector is capable of extracting vulnerability-related information from Twitter, augment it with additional sources and predict if the vulnerability is exploitable in a real-world scenario.
An interesting feature from this work is the consideration of adversarial interference to deceive the classifier. 
%The authors developed a threat model with three types of adversaries. 
These adversaries vary in their knowledge of the classifier and in the complexity of their poison-attacks that attempt to deceive the classifier.
Currently, our system uses sets of pre-defined user accounts based on the likelihood that they tweet about the security of the protected IT infrastructure. 
%We have not yet implemented a reputation system capable of fetching new accounts, however the existence of adversarial agents should be kept in mind.

Regarding IoC extraction from OSINT, Liao et al.~\cite{paper_IOC_game} developed iACE, an automated tool capable of IoC extraction.
It extracts information from technical articles, which tend to provide information in a more structured and formal manner when compared to Twitter.
The authors take advantage of some properties of these articles, such as a set of context terms and their grammatical relations.
Unlike iACE, our approach aims to explore the use of deep learning techniques to extract grammatical relations automatically, removing the necessity for manual feature engineering.

Concurrently with this work, Zhou et al.~\cite{Zhou} employed a NER architecture for extracting IoC from cybersecurity reports. Differently from our architecture, their proposal integrates manually engineered features and an attention mechanism with the BiLSTM. 

\subsection{Deep learning for tweet analysis}
Using a similar architecture to the one we use, Severyn et al.~\cite{Twitter_sentiment_1} perform sentiment analysis on Twitter data taken from the SemEval 2015 Twitter Sentiment Analysis challenge.
In comparison to other systems submitted to the challenge, their solution ranked in the two first positions in both tasks.

Badjatiya et al. \cite{Twitter_hate} experimented with several classifiers including, but not limited to, deep learning methods to detect hate speech in tweets.
The authors report that deep learning techniques perform significantly better than other methods.

Regarding applications of deep learning for NER tasks using Twitter data, Jimeno-Yepes et al. \cite{JimenoYepes2016NERFM} implemented a sequence-to-sequence LSTM architecture for the annotation of medical entities to support public health surveillance.
The architecture presented outperforms the previous state-of-the-art.
Regarding future work, the authors plan on using the architecture proposed by Lample et al. \cite{paper_BiLSTMCRF}, which is the one in which we have based our NER approach.

\section{Tool Architecture}

% A figura ou fica na primeira pagina ou vai para a Secção III onde é mencionada
\begin{figure}[!t]
\centering
\includegraphics[width=\linewidth]{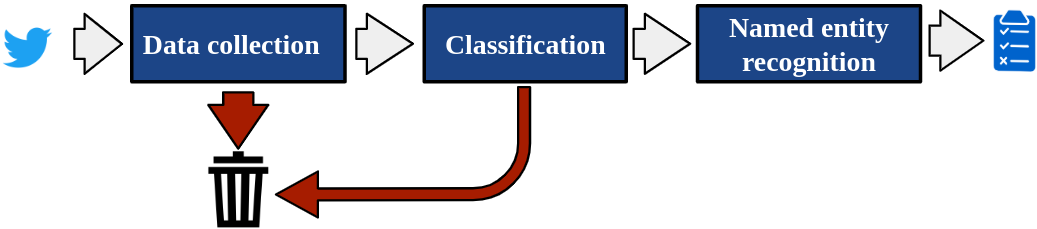}
\caption{Twitter threat detection pipeline.}
\label{fig:og_pipeline}
\end{figure}

Our tool follows the high-level three-stage pipeline architecture depicted in Figure~\ref{fig:og_pipeline} to process data from Twitter and extract relevant cybersecurity information.
The first stage collects tweets through the Twitter API, filters them based on a set of keywords and normalizes tweets to a specific format.
In the second stage, a binary classifier labels tweets as either relevant, meaning the tweets are likely to contain valuable information about an asset, or irrelevant otherwise.
Finally, in the information extraction stage, relevant tweets are processed by an NER network.
The information extracted can be used to issue a security alert or to enrich an existing IoC in a threat intelligence platform such as MISP~\cite{paper_MISP}.
The following sections describe the three stages.

 \subsection{Data collection}

The data collection stage comprises the execution of three tasks: data collection, keyword-based pre-filtering, and text pre-processing.

\subsubsection{Collector}
The collector requires a set of Twitter accounts from which it will query the Twitter's streaming API.
These accounts can be from companies, security vendors, hacker groups, security analysts, and researchers.
Ideally, they are chosen based on their likelihood of tweeting about components present in the monitored IT infrastructure.
This approach of retrieving tweets from a specific set of accounts can drastically reduce the amount of irrelevant information that is gathered by the collector.

\subsubsection{Filter}
By assuming that a tweet about an infrastructure asset has to mention its properties and components, the filter module employs a set of user-defined keywords describing the infrastructure being monitored to drop irrelevant tweets, thus further decreasing the amount of information that flows through the pipeline.
For instance, if an analyst desires to be informed about potential threats to a web service hosted on a cloud platform, the set of keywords should include operating systems, the cloud platform being used and all other components supporting the asset in question.

\subsubsection{Pre-processing}
At the end of the data collection stage, a pre-processing module transforms tweets that passed the filter, making their representation uniform before subsequent processing by the classifier and NER modules.
Characters are converted to lower-case, and hyperlinks and special characters are removed, except for `.', `-', `\_', and `:', as they are often used in IDs, version numbers, or component names.

\subsection{Classification}
\begin{figure}[!t]
\centering
\includegraphics[width=\linewidth]{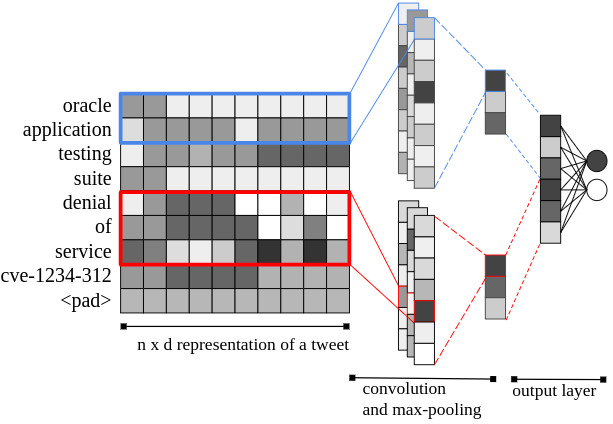}
\caption{Convolutional Neural Network architecture for sentence classification.}
\label{fig:cnn}
\end{figure}

The classification stage aims to detect tweets that contain security-related information so that only these will proceed to the Named Entity Recognition (NER) stage.
%The classification stage aims to detect tweets that contain security-related information, and these will be sent to the Named Entity Recognition stage.
To perform this task, we implemented a binary classifier using a CNN~\cite{paper_kim} whose architecture can be described by five layers: input, embedding, convolution, max-over-time-pooling, and output.
Figure~\ref{fig:cnn} shows this architecture. 
 
 \subsubsection{Input layer}
The CNN receives a sentence represented by a sequence of $n$ integers, each representing a word token, just as shown in Figure~\ref{fig:cnn}.

\subsubsection{Embedding layer}
Each integer corresponds to a $d$-dimensional numeric vector representing the semantic value of the corresponding word.
These \emph{word vectors} can be randomly initialized or extracted from previously trained language models (e.g., GloVE~\cite{paper_Glove} or Word2Vec~\cite{paper_word2vec}) providing a starting point for the semantic value.
In both cases, the learning algorithm may further adjust the word vector.
As a result of this layer, we have an $n \times d$ sentence embedding matrix $\bm{S}$.

 \subsubsection{Convolution layer}
With a set of $k$ learnable  kernels, each containing $f$ filters with height $h$ and width $d$, the convolution operation will slide down these kernels over the embedded matrix $\bm{S}$, producing $k \times f$ feature maps.
The height and width of the filters correspond to the number of words covered and to the dimension of the word vectors, respectively.
As the filter slides down the matrix, the resulting feature map will have a length of $n - h + 1$, where $n$ is the number of words in the sentence.

A single feature $c_i$, produced by filter $\bm{W}$, is computed as,
\begin{equation}
 c_i = \sigma(\bm{W} \odot \bm{S}_{i:i+h-1} + b)\,,
\end{equation}
where $\sigma$ is a non-linear function, $\odot$ denotes the Hadamard product, $b$ is a bias term, and $\bm{S}_{i:i+h-1}$ denotes a sub-matrix from $\bm{S}$ by taking rows $i$ to $i+h-1$.
Given that this operation computes one feature for each stride of the filter over the embedded matrix, the convolutional layer outputs a set of $k \times f$ feature map vectors, each containing $n-h+1$ features.

%\textbf{TODO: nao deviamos citar de onde essas ideias vem? Ha algo novo aqui alem do que se faz normalmente nessa layer? 
%ANSWER: já referimos acima que é baseada no Kim~\cite{paper_kim} e até ao momento fazemos nada extra}

\subsubsection{Max-over-time-pooling layer}
Intuitively, if one feature map results from sliding a filter over the embedding matrix, the feature values composing that map denote how strong the feature is within a specific input window.
We reduce such feature map into a single value providing information about the presence or absence of a feature, through the max-pooling operation~\cite{paper_Collobert_scratch}.
In the end, this layer helps to filter redundant information, thus preventing overfitting and decreasing the computational burden.
As each feature map gets reduced to a single value, this layer outputs a $k \times f$ feature vector.

\subsubsection{Output layer}
Before using the selected features in the output layer of the network, dropout~\cite{Dropout} is applied to the feature vector. 
This procedure sets a proportion of the feature vector elements to zero, preventing gradients to propagate through these elements and co-adapting the nodes.
Dropout acts as a form of regularization~\cite{Wag13}, preventing overfitting and promoting generalization.
Finally, feature nodes are used by a fully-connected softmax layer which outputs the probability of a tweet to contain relevant information.
If relevant, the tweet moves to the next stage of the pipeline, otherwise it is dropped.

%\textbf{TODO: Nao devia ter uma referencia logo apos o softmax?
%ANSWER: Acho que não, sendo que o tal softmax layer é algo que já é usual de redes neuronais, e ainda não vi %ninguem a citar algo para essa camada..}

\subsection{Named entity recognition}
\begin{figure}[t]
\centering
\includegraphics[width=\linewidth]{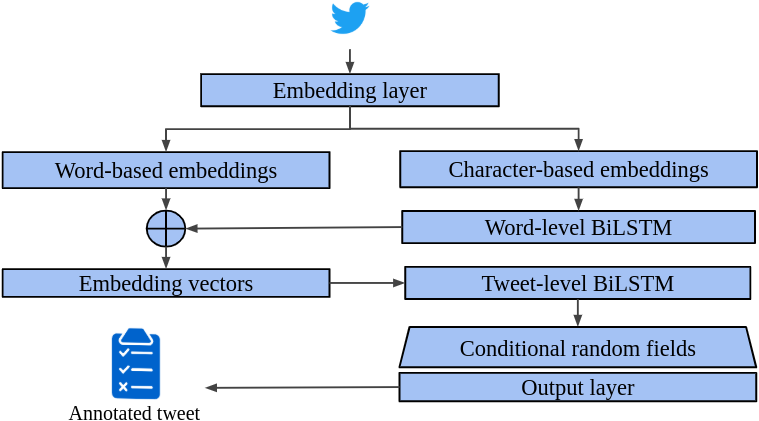}
\caption{Bidirectional Long Short-Term Memory architecture for named entity recognition.}
\label{fig:ner}
\end{figure}
In the NER phase, we aim to extract information from tweets that have been considered relevant by the classifier.
We have based our model on a BiLSTM neural network~\cite{paper_BiLSTMCRF}.
Figure~\ref{fig:ner} depicts its architecture.

This network locates and labels valuable security-related entities such as monitored infrastructure assets, vulnerabilities, attacks, and vulnerability repository IDs mentioned in tweets.
To the extent possible, we defined security-related entities according to descriptions from the ENISA risk management glossary~\cite{ENISA}.
In the following, we briefly describe the network layers: input, embedding, word-level BiLSTM, tweet-level BiLSTM, and output.

\subsubsection{Input layer}
In addition to the $n$ integers representing the word tokens applied to the input layer of the classifier, this network receives secondary sequences of $n_c$ integers, one for each word at the input.
Each secondary sequence represents the characters that form a word token.
These sequences allow the network to be less constrained by the vocabulary used during training.

\subsubsection{Embedding layer}
This layer's functionality is similar to the classifier's regarding the sequence of $n$ integers that correspond to the words in tweets.
These are converted to a $d_w$-dimensional numeric vector, thus providing a word-level semantic representation of the tweet.
This results in an $n \times d_w$ matrix representation where $n$ is the number of words and $d_w$ is the length of the embedded word vectors.

Additionally, the $n_c$ integers in each secondary sequence are converted to $d_c$-dimensional numeric vectors, providing a character-level representation of the tweet.
Therefore, besides the word-level embedded matrix, each word has a corresponding $c \times d_c$ character-level matrix representation, where $c$ is the number of characters in a word and $d_c$ is the length of the embedded character vectors.

\subsubsection{Word-level BiLSTM layer}
The embedded character matrix corresponding to each word is fed to a BiLSTM network, containing two cells that read the sequence of input character vectors in opposite directions.
Both cells possess an $h_c$-dimensional vector hidden state, that is updated at every time step (i.e., at every character read).
After reading all the characters in the embedded character matrix, the cell states are extracted and concatenated.
These vectors hold a character-level representation from both left-to-right and right-to-left readings.

As a result, this layer generates an $n \times (2 \times h_c)$ state matrix, for concatenation with the word-level embedding matrix.

\subsubsection{Tweet-level BiLSTM layer}
At this stage we have a matrix of $n$ words, each word represented by a numeric vector with dimension $d_w + (2 \times h_c)$.

Similar to the process described for the word-level BiLSTM, we feed this tweet representation to another BiLSTM layer, word by word.
However, while in the previous layer we only retrieve the final hidden state, in this layer we read it at every timestep (every word representation read).

Thus, the output of this layer is an $n \times h_t$ matrix, where $h_t$ is the length of the BiLSTM's cell state vector.

\subsubsection{Output layer}
In the final layer of the NER model, we have a fully-connected neural network and a Conditional Random Fields (CRF) module \cite{paper_CRF}.
The fully-connected neural network contains $m$ neurons, $m$ being the number of possible labels that the model can predict.
As an input to this layer, we have the $n \times h_t$ matrix provided by the previous layer, therefore creating an $n \times m$ output matrix.
This matrix can be thought as a score matrix, where the activations of the  $m$ neurons produce a score for each label.

Then, this scoring matrix is used to compute the final sequence of labels.
Instead of choosing the highest label score for each word, thus modeling labeling decisions independently, the CRF module allows to model them jointly.
Given a sequence of words and corresponding score vectors, and candidate label sequences, a linear-chain CRF computes a single score for each candidate sequence of labels.
By selecting the highest single score, this layer outputs the corresponding candidate sequence of $n$ labels that match the input sequence of $n$ words at the input layer.

\section{Experimental Setup}
This section describes the experimental work conducted to design both the classification and NER neural networks.
Additionally, we describe the approach employed to obtain results for other state-of-the-art or well-established methodologies, for comparison purposes.

\subsection{Datasets}
To build our system, we continuously collect tweets from two sets of accounts, denoted {S1} and {S2}.
The tweets are related to IT infrastructures specified by three private organizations:  a worldwide travel services provider, a global-company cybersecurity department, and a nation-wide power utility.
Each of the three infrastructures denoted {A}, {B}, and {C}, is defined by a set of assets, which in turn are specified by a variable number of keywords.
For a subset of tweets collected over four-months, tweets were analyzed and labelled relevant (or not) for the security of the designated assets.\footnote{https://github.com/ndionysus/twitter-cyberthreat-detection}

Each infrastructure dataset was split and organized into three subset groups, as described in Table \ref{table:casestudy_data}, one for training the models ({A1}, {B1}, {C1}), another for validation ({A2}, {B2}, {C2}) and the remaining one for testing ({A3}, {B3}, {C3}).
% A == Atos
% B == Amadeus
% C == EDP
\setlength\cellspacetoplimit{2pt}
\begin{table}[t!]
\centering
\caption{Datasets corresponding to companies A, B and C.}
\resizebox{\linewidth}{!}{
\begin{tabular}{|Sc|Sc|Sc|Sc|Sc|Sc|}
\hline
{Datasets} & {Time Interval} & {Accounts}  & {Positives} & {Negatives} & {Total} \\ \hline \hline
{A1} & \multirow{3}{*}{\parbox{1.5cm}{\centering 21/11/2016\\to\\27/01/2017}} & \multirow{6}{*}{S1} & 1074  & 694 & 1768  \\ \cline{1-1} \cline{4-6} 
{B1} & & & 1201  & 638 & 1839  \\ \cline{1-1} \cline{4-6} 
{C1} & & & 1293  & 1473  & 2766  \\ \cline{1-2} \cline{4-6} 
{A2} & \multirow{3}{*}{\parbox{1.5cm}{\centering 27/01/2017\\to\\27/02/2017}} & & 282 & 767 & 1049  \\ \cline{1-1} \cline{4-6} 
{B2} & & & 387 & 671 & 1058  \\ \cline{1-1} \cline{4-6} 
{C2} & & & 325 & 592 & 917  \\ \hline \hline
{A3} & \multirow{3}{*}{\parbox{1.5cm}{\centering 27/02/2017\\to\\27/03/2017}} & \multirow{3}{*}{S1 + S2} & 219 & 313 & 532  \\ \cline{1-1} \cline{4-6} 
{B3} & & & 250 & 247 & 497  \\ \cline{1-1} \cline{4-6} 
{C3} & & & 289 & 358 & 647  \\ \hline
\end{tabular}
}
\label{table:casestudy_data}
\end{table}

The validation subset is distinguished solely by the time interval, being collected one month after the training subset.
More importantly, the testing subset is distinguished from the others not only by the time interval but also because it includes tweets from an additional set of accounts.
We have chosen this strategy so that the models' performance and generalization capability assessment considers both tweets in the future of training data and a combination of that with previously unseen Twitter accounts.

Table \ref{table:tags} displays the labels considered for the NER task.
Although we defined the entities considering descriptions from the ENISA risk management glossary~\cite{ENISA}, we adopted broad definitions, so that the labels become more distinguishable by the network.
For example, the label \texttt{VUL} includes both vulnerabilities and threats.
The reason for this is twofold.
On the one hand, we have limited data for training.
Thus, more specific entity definitions could result in too few instances in the datasets.
On the other hand, as Twitter is an informal OSINT source, many users of interest do not follow any international standard when referring to security matters.
Once more data is available and labeled, the number of entities could be extended, allowing for a more accurate taxonomy.
%===================== TABELA DE TAGS
\setlength\cellspacetoplimit{2pt}
\begin{table}
\caption{Named entities to be extracted from a tweet.}
\begin{tabularx}{\linewidth}{|Sc|X|}
\hline
{Label}   & {Description}  \\ \hline \hline
  O   	& Does not contain useful information.  \\ \hline
  ORG & Company or organization. \\ \hline
  PRO & A product or asset. \\ \hline
  VER & A version number, possibly from the identified asset or product. \\ \hline
  VUL & May be referencing the existence of a threat or a vulnerability. \\ \hline
  ID & An identifier, either from a public repository such as the National Vulnerability Database (NVD)~\cite{NVD}, or from an update or patch. \\ \hline
\end{tabularx}
\label{table:tags}
\end{table}

\setlength\cellspacetoplimit{2pt}
\begin{table*}[t!]
\caption{Binary classifier results across all infrastructures.}
\resizebox{\linewidth}{!}{%
\begin{tabular}{Sc|Sc|Sc|Sc|Sc|Sc|Sc|Sc|Sc|Sc|Sc|Sc|Sc|Sc|Sc|Sc|Sc|Sc|Sc|}
\cline{2-19}
 & \multicolumn{2}{c|}{{A1}} & \multicolumn{2}{c|}{{A2}} & \multicolumn{2}{c|}{{A3}} & \multicolumn{2}{c|}{{B1}} & \multicolumn{2}{c|}{{B2}}  & \multicolumn{2}{c|}{{B3}}   & \multicolumn{2}{c|}{{C1}} & \multicolumn{2}{c|}{{C2}}   & \multicolumn{2}{c|}{{C3}}   \\ \cline{2-19} 
 & \multicolumn{1}{c|}{TPR} & \multicolumn{1}{c|}{TNR} & TPR & TNR   & TPR & TNR   & TPR & TNR   & TPR   & TNR   & TPR   & TNR   & TPR & TNR   & TPR   & TNR   & TPR   & TNR   \\ \hline
\multicolumn{1}{|c|}{SVM} & 
% A
0.953 & 0.970 & 0.915  & 0.866 & 0.785  & 0.831 &
% B
0.977 & 0.943 & 0.995 & 0.450 & 0.996 & 0.575 &
% C
0.968 & 0.972 & 0.951 & 0.586 & 0.989 & 0.555 \\ \hline
\multicolumn{1}{|c|}{MLP} &
% A
0.976 & 0.987 & 0.867  & 0.951 & 0.739  & 0.926 &
% B
0.992 & 0.977 & 0.858 & 0.910 & 0.818 & 0.882 &
% C
0.993 & 0.992 & 0.843 & 0.878 & 0.888 & 0.862 \\ \hline \hline
\multicolumn{1}{|c|}{CNN + Random} & 
% A
0.988 & 0.995 & 0.916  & 0.904 & 0.810  & 0.886 & 
% B
\textbf{0.998} & \textbf{0.992} & 0.918 & 0.891 & 0.925 & 0.890 &
% C
0.989 & 0.995 & 0.873 & 0.871 & 0.954 & 0.901 \\ \hline
\multicolumn{1}{|c|}{CNN + GloVE} & 
% A
0.979 & 0.979 & 0.936  & 0.919 & 0.839  & 0.901 &
% B
0.993 & 0.995 & \textbf{0.959} &  \textbf{0.930}&  \textbf{0.956}&  \textbf{0.927}&
% C
0.993 & 0.993 & 0.890 & 0.864 & 0.966 & 0.904 \\ \hline

\multicolumn{1}{|c|}{CNN + Word2Vec} & 
% A
\textbf{0.998} & \textbf{0.999} & \textbf{0.943}  & \textbf{0.932} & 0.849  & 0.927 &
% B
0.994 & 0.979 & 0.949 & 0.911 & 0.947 & 0.915 &
% C
\textbf{0.995} & \textbf{0.999} &  \textbf{0.917}&  \textbf{0.875}&  \textbf{0.958}&  \textbf{0.911}\\ \hline
\multicolumn{1}{|c|}{BiLSTM + Random} &
% A
0.999 & 0.971 & 0.915  & 0.906 & \textbf{0.913}  & \textbf{0.904} &
% B
0.978 & 0.978 & 0.920 & 0.949 & 0.912 & 0.964 &
% C
0.990 & 0.932 & 0.865 & 0.872 & 0.938 & 0.911 \\ \hline
\multicolumn{1}{|c|}{BiLSTM + GloVE} &
% A
0.986 & 0.979 & 0.897  & 0.940 & 0.863  & 0.946 &
% B
0.973 & 0.983 & 0.951 & 0.918 & 0.936 & 0.943 &
% C
0.985 & 0.975 & 0.874 & 0.870 & 0.955 & 0.911 \\ \hline
\multicolumn{1}{|c|}{BiLSTM + Word2Vec} & 
% A
0.995 & 0.995 & 0.901  & 0.897 & 0.890  & 0.911 &
% B
0.977 & 0.985 & 0.907 & 0.951 & 0.912 & 0.951 &
% C
0.997 & 0.989 & 0.880 & 0.872 & 0.934 & 0.919 \\ \hline
\end{tabular}
\label{table:svmvscnn}
}
\end{table*}

\subsection{Training and evaluation methodology} 
\label{Training_Eval_method}
To compare the deep-learning neural network classifier methodology described in previous sections we used a linear approach, the Support Vector Machine (SVM)~\cite{paper_SVM}, and a non-linear one, the Multi-Layer Perceptron (MLP) \cite{rosenblatt1958perceptron,rumelhart1985learning}.

The described deep-learning NER methodology was compared to the Stanford CRF NER approach~\cite{StanfordNER}, trained with the procedure and parameters suggested by the authors.

The two deep architecture models described in previous sections were implemented using TensorFlow~\cite{paper_tensorflow}.
Training was conducted with the TensorFlow implementation of the Adam optimizer ~\cite{paper_Adam} using the default parameters.
The procedure was executed for 100 epochs with a batch size of 256, using the validation dataset for early stopping.
The SVM and MLP models were implemented using the Apache Spark Machine Learning library~\cite{meng2016mllib}.
Training was carried out using the Stochastic Gradient Descent (SGD)~\cite{bottou2010large} optimizer for the SVM model, and the Limited-memory  Broyden–Fletcher–Goldfarb–Shanno (L-BFGS)~\cite{liu1989limited} optimizer for the MLP.
In both cases, most parameters have been kept equal to the package defaults, except those considered in a grid search executed to select appropriate models.
For the SVM, 100 iterations were employed, whereas 200 were used for the MLP.

As a grid search was used to select appropriate hyper-parameter values and design variables settings for these models, in all cases a 10-fold cross-validation methodology was employed.
The 10-fold results were averaged and used to select the best models and variants for comparison.
Regarding the classification task, besides using the CNN described, we also adapted the BiLSTM architecture for classification.
The adaptation consists of removing the CRF layer, using only the last state of both LSTMs, and having only two neurons at the output for binary classification.

Having picked the best configuration of each model type and variants for each infrastructure, we train the models with the full training set, using the validation sets for early stopping, and then test on the testing sets (A3, B3, and C3).

Next, we describe the grid search procedure executed, highlighting the hyper-parameter values and design variable settings considered.

\subsection{Model optimization using grid search}
\label{grid_search}
\subsubsection{Embedding Vectors and BiLSTM hidden states}
Regarding the initialization of the embedding vectors of deep neural architectures, we tested randomly initialized embedding, GloVE pre-trained embedding~\cite{paper_Glove}, and Word2vec pre-trained embedding~\cite{paper_word2vec}.
For the random initialization case, we tested with vector dimensions of 100, 200, and 300.
The vector dimension applied in pre-trained embedding approaches was 300.
Regarding the NER BiLSTM approach, character embedding was done only using random initialization with vector dimension varied within $\{25, 50, 100\}$.

For the NER model, the sizes of both state vectors varied within the same alternatives as the randomly initialized embedding vectors.

\subsubsection{CNN Convolutional layer}
In the convolutional layer of the CNN, we varied the kernels in number, height, and number of filters, int the following way:
\begin{itemize}
 \item {Number of kernels:} varied between 3 and 6;
 \item {Kernel heights:} varied incrementally according to the number of kernels, either in a sequential manner (e.g. 2,3,4) or by parity (e.g., 3,5,7, or 2,4,6);
 \item {Number of filters}: varied within $\{64, 128, 192, 256\}$;
\end{itemize}

\subsubsection{Dropout regularization}
Considering the CNN classification model, the dropout rate varied within $\{0.3, 0.5, 0.6\}$.
The NER BiLSTM models used two dropout layers, the first after the word-level BiLSTM and the second after the sentence-level BiLSTM.
We considered only two settings: no dropout or 0.5 dropout probability, which resulted in four possible variants.

\subsubsection{SVM and MLP models}
The SVM and MLP classifiers use the Term Frequency - Inverse Document Frequency (TF-IDF) approach to obtain a numeric representation of tweets.
The TF-IDF feature vector size was varied within the set:$\,\{30, 50, 80, 100, 200, 300, 500, 750, 1000, 1500, 3000\}$.
For the SVM, the regularization parameter \texttt{C} was varied within $\{0.01, 0.02, 0.05, 0.1, 0.2, 0.5, 1, 2, 5\}$, while the SGD step size was varied within $\{0.1, 0.5, 1, 1.5, 2, 5\}$.
Regarding the MLP, the number of layers varied from 2 to 8 and the number of neurons within $\{5, 7, 10, 12, 14, 16, 18, 20\}$.

\section{Results}
Regarding the tweet classification models, we used the True Positive Rate (TPR) and True Negative Rate (TNR) metrics for evaluation, as these carry useful information for SOC analysts.
For the NER task, the models were evaluated using the precision, recall, and F1 metrics.

The results presented shows the best model found in the grid search executed for each model type and variant.
In the classification task, we consider best the model whose (TPR, TNR) is closest in the Euclidean sense to the (1, 1) ideal result.
For NER, we adopted the best F1-score to find the best model.

\subsection{Tweet classification}
Table \ref{table:svmvscnn} summarizes the results obtained by the different classification models tested, highlighting the different word embedding approaches.
The result pairs in boldface denote the best compromise between TPR and TNR in each dataset.

The results indicate that across the three infrastructures the CNN achieves the best overall results and pre-trained embedding vectors are preferable to randomly initialized ones.
The exception occurs in infrastructure A testing set, in which case a BiLSTM model with randomly initialized embedding vectors achieved the best result.
Well-established techniques such as SVMs or MLPs produced worse results, being incapable of achieving a good balance over TPR and TNR in several cases.
Using frequency-based features instead of semantic word vectors may have
influenced the results obtained.
We tested both SVM and MLP models with concatenated pre-trained
word-vector inputs.
However, this approach offered poor results, therefore it was not
further explored.
Regarding generalization, we observe that the performance degradation is small, leaving the results comfortably above 90\% in most cases.
Importantly, we conclude that there is no significant performance degradation resulting from the addition of tweets from a second set of accounts in the testing sets.

Focusing on the CNN results, Table \ref{table:archsallcnn} displays the best configurations that resulted from the grid search for each infrastructure.
\setlength\cellspacetoplimit{2pt}
\begin{table}[!t]
\centering
\caption{Architectures of the best performing models for each dataset.}
%\resizebox{\linewidth}{!}{%
\begin{tabular}{|Sc|Sc|Sc|Sc|Sc|Sc|}
\hline
{Dataset} & {Kernels} & {Number of Filters} & {Dropout Rate}\\ \hline % & \textbf{Embedding Model} \\ \hline
{A} & {[}3,5,7{]} & 128 & 0.5\\ \hline % & Word2Vec \\ \hline
{B} & {[}3,5,7{]} & 256 & 0.5\\ \hline % & GloVE \\ \hline
{C} & {[}3,5,7,9{]} & 256 & 0.5\\ \hline % & Word2Vec \\ \hline
\end{tabular}%
%}
\label{table:archsallcnn}
\end{table}
The results favour models with a small number of kernels with oddly spaced heights ($[3,5,7]$ means three filters with heights 3, 5, and 7), a large number of filters and 0.5 dropout probability.

% As tabelas individuais estão no old.tex

%%%%EXPERIMENTAL
\setlength\cellspacetoplimit{1pt}
\begin{table*}[t]
\centering
\caption{NER F1-score results across all infrastructures.}
\begin{tabular}{Sc|Sc|Sc|Sc|Sc|Sc|Sc|Sc|Sc|Sc|}
\cline{2-10}
 & {A1} & {A2} & {A3} & {B1} & {B2} & {B3} & {C1} & {C2} & {C3} \\ \hline
\multicolumn{1}{|c|}{{Stanford CRF}} & 
\textbf{1.000}  & 0.917 & 0.852 & \textbf{0.999}  & 0.879 & 0.810 & \textbf{0.999}  & 0.899 & 0.906 \\ \hline \hline
\multicolumn{1}{|c|}{{BiLSTM + Random}}   & 
0.999  & \textbf{0.932} & \textbf{0.906} & \textbf{0.999}  & 0.919 & 0.882 & 0.987  & 0.925 & 0.928 \\ \hline
\multicolumn{1}{|c|}{{BiLSTM + GloVE}} & 
0.999  & 0.926 & 0.894 & 0.997  & \textbf{0.932} & 0.888 & 0.998  & 0.924 & 0.932 \\ \hline
\multicolumn{1}{|c|}{{BiLSTM + Word2Vec}} & 
\textbf{1.000}  & 0.893 & 0.864 & 0.998  & 0.927 & \textbf{0.890} & 0.984  & \textbf{0.928} & \textbf{0.934} \\ \hline
\end{tabular}
\label{NER_results_ALL}
\end{table*}
%%%%

\setlength\cellspacetoplimit{4pt}
%%%%%%%%%%%%%
%TABELA NVD VS TWITTER
%%%%%%%%%%%%%
\begin{table*}[h]
\caption{Vulnerabilities published on twitter prior to being disclosed in NVD}
\label{table:NVDvsTwitter}
\resizebox{\linewidth}{!}{%
\begin{tabular}{Sc|Sc|Sc|Sc|Sc|}
\cline{2-5}
& {NVD date}   & {Tweet date} & {Tweet} & {CVSS} \\ \hline
\multicolumn{1}{|l|}{{A2}}  & \multicolumn{1}{l|}{06/02/2017} & \multicolumn{1}{l|}{31/01/2017} & 
$\text{Vuln:}~\ner{PRO}{Linux Kernel}~\ner{ID}{CVE-2017-5546}~\ner{VUL}{Local Denial of Service Vulnerability}~\text{https://t.co/bLEJIb1ZVD}$ & 7.8\\ \hline
\multicolumn{1}{|l|}{\multirow{3}{*}{{A3}}} & 08/06/2017 & 27/03/2017 &
$\ner{PRO}{VMware Player} \ner{VER}{12.x \&lt 12.5.4} \text{Drag-and-Drop Feature Guest-to-Host} \ner{VUL}{Code Execution} \ner{ID}{(VMSA-2017-0005)} \ner{PRO}{(Linux)} \text{https://t.co/xMIP5JlvOZ}$ & 9.9
\\ \cline{2-5} 
\multicolumn{1}{|Sc|}{} & 27/03/2017 & 24/03/2017 & 
$\text{Vuln:}~\ner{PRO}{Broadcom}~\ner{ID}{BCM4339}~\text{SoC}~\ner{ID}{CVE-2017-6957}~\ner{VUL}{Stack-Based Buffer Overflow Vulnerability}~\text{https://t.co/vR6EznOsBi}$ & 8.1 
\\ \hline
\multicolumn{1}{|Sc|}{{B2}}  & 20/03/2017 & \multicolumn{1}{l|}{29/01/2017} & 
$\text{Vuln:}~\ner{PRO}{Apache Tomcat}~\ner{ID}{CVE-2016-6816}~\ner{VUL}{Security Bypass Vulnerability}~\text{https://t.co/PfOdfDGIfy}$ & 7.1 \\ \hline
\multicolumn{1}{|l|}{\multirow{3}{*}{{B3}}} & 27/07/2018 & 01/03/2017 & 
$\text{Vuln:}~\ner{PRO}{Red Hat}~\text{CloudForms Management Engine}~\ner{ID}{CVE-2017-2632}~\ner{VUL}{Privilege Escalation Vulnerability}~\text{https://t.co/Vm0fMMM1Rc}$ & 4.9 
\\ \cline{2-5} 
\multicolumn{1}{|Sc|}{} & 20/03/2017 & 16/03/2017 & 
$\text{Vuln:}~\ner{PRO}{Apache Tomcat}~\ner{ID}{CVE-2016-6816}~\ner{VUL}{Security Bypass Vulnerability}~\text{https://t.co/FK5nXKcfy8 \#bugtraq}$ & 7.1 
\\ \hline
\multicolumn{1}{|l|}{\multirow{3}{*}{{C2}}} & 16/03/2017 & 06/02/2017 & 
$\text{\#Vuln:}~\ner{ORG}{\#Microsoft}~\ner{PRO}{\#Windows}~\ner{ID}{CVE-2017-0016}~\ner{VUL}{Memory Corruption Vulnerability}~\text{https://t.co/ZR3DVVgx3j \#bugtraq}$ & 5.9 
\\ \cline{2-5} 
\multicolumn{1}{|Sc|}{} & 15/02/2017 & 14/02/2017 & 
$\ner{ID}{ZDI-17-109}~\text{:}~\ner{PRO}{Adobe Flash Player}~\text{MessageChannel}~\ner{VUL}{Type Confusion Remote Code Execution Vulnerability}~\text{https://t.co/hTaiCS671W}$ & 8.8
\\ \hline
\multicolumn{1}{|l|}{\multirow{3}{*}{{C3}}} & 27/07/2018 & 01/03/2017 & 
$\text{Vuln:}~\ner{PRO}{Red Hat}~\text{CloudForms Management Engine}~\ner{ID}{CVE-2017-2632}~\ner{VUL}{Privilege Escalation Vulnerability}~\text{https://t.co/Vm0fMMM1Rc}$ & 4.9 
\\ \cline{2-5}
% FALSO POSITIVO, SERÁ QUE DEVIA SER INSERIDO?
%\multicolumn{1}{|l|}{} & 15/12/2017 & 08/03/2017 & 
%$\text{Vuln:}~\ner{PRO}{Flash}~\text{Seats for iOS}~\ner{ID}{CVE-2017-3190}~\ner{VUL}{SSL %Certificate Validation Security Bypass Vulnerability}~\text{https://t.co/xi9sdB2yGt}$& 7.5 
%\\ \cline{2-5} 
\multicolumn{1}{|l|}{} & 11/06/2018 & 08/03/2017 & 
$\text{\#Vuln:}~\ner{ORG}{\#Mozilla}~\ner{PRO}{\#Firefox}~\ner{ID}{MFSA 2017-05}~\ner{VUL}{Multiple Security Vulnerabilities}~\text{https://t.co/POFeaWjREj \#bugtraq}$ & 7.5
\\ \hline
\end{tabular}
}
\end{table*}

\subsection{Named entity recognition}
% Tirei o Stanford's e voltei a meter Stanford porque o software chama-se mesmo Stanford CRF NER
Table \ref{NER_results_ALL} displays the best performing models, including the results obtained by the Stanford CRF NER approach.

In comparison to the Stanford CRF, in general, the BiLSTM architectures achieved better results across all infrastructures.
The exception is the Precision metric achieved in datasets A2 and C2, where the Stanford CRF showed equal performance.
In general, BiLSTM networks correctly detect the target entities in more than 90\% of the cases.

Except for infrastructure A, the use of pre-trained embedding word vectors provided a slight gain.
However, the improvement is not significant enough to provide a definitive conclusion.
Unlike in the tweet classification task, where both security-related and non-security-related tweets are present, in the NER task, only the former is present.
Thus, the advantage of using pre-trained embedding vectors based on non-security related text may not be as pronounced.

Although in infrastructures \texttt{A} and \texttt{C} the best models obtained the highest F1-score in both validation and testing sets, in the case of infrastructure \texttt{B} a model different from the best performing model of the validation set obtained a better F1-score.
However, the difference between these is negligible.
Thus, the choice of the best model may depend on favoring the testing set F1-score over the validation sets' or on considering an average of both preferable to make the decision.

%Regarding the configuration of the BiLSTM models, for the dimension of the character embedding vector and the character-level hidden state, the best performing models favored (8 out of 9) the highest available value of $100$.
Regarding the dimension of the character embedding vector and the character-level hidden state of BiLSTM networks, the best performing models favored (8 out of 9) the highest available value of $100$.
%Considering the word embedding vector dimension, the best models used mostly (7 out of 9) a 300 dimension.
% 6 deles apenas tiveram a opcao de 300 porque são pre-treinados
For the word-level hidden state vector dimension, most models (8 out of 9) used the smallest available value of $100$.
Finally, regarding the dropout layers, no clear trend could be observed in the results.

% A tabela dos hiperparametros da BiLSTM esta no old.tex

\section{Analysis of Indicators of Compromise}
Regarding the applicability of Twitter for cyberthreat awareness, we analyzed the tweets labeled relevant by the classifier in the validation and testing sets.
By using the \texttt{ID} label, we analyzed the corresponding NVD vulnerability entry to verify the existence of tweets mentioning these vulnerabilities priorly to the disclosure date, and to find their severity according to the Common Vulnerability Scoring System (CVSS)~\cite{CVSS}.

A sample of such tweets found is displayed in Table \ref{table:NVDvsTwitter}.
Each entry shows the tweet date and the NVD disclosure date, the tweet with the labels identified by the NER model, and the CVSS score.
The number of days since tweet publication to NVD disclosure ranged from 1 to 148, clearly showing the timeliness with which the deep neural network-based OSINT processing pipeline can provide vulnerability information to organizations' SOCs. 
The tweets CVSS score range from a medium 4.9 score to a 9.9 critical score, thus showing the relevance of the information found.

The BiLSTM NER model recognized the most important aspects of these tweets, such as the infrastructure asset, the vulnerability, and useful identifiers such as the Common Vulnerabilities and Exposures (CVE).
These identified entities could then have been used to issue a security warning or to fill an IoC in a threat sharing platform.

Although our current datasets are not large, the results obtained and the information relevance and timeliness justify the possibility of using Twitter as an OSINT source for cyberthreat discovery.
Furthermore, even though we did not identify one case in the datasets used where the tweet references a zero-day exploit without mentioning a CVE or similar identifier, such was the case in late August of 2018.
A Twitter user made public a zero-day vulnerability in Microsoft Windows' task scheduler, providing a proof-of-concept exploit.\footnote{https://www.zdnet.com/article/windows-zero-day-vulnerability-disclosed-through-twitter/}
We sent the original tweet through our pipeline and the classifier and NER models correctly labeled the tweet as relevant and identified the asset in question to be Microsoft Windows.
The exploit was officially made public only on the 9\textsuperscript{th} of September, regardless of the original tweet appearance on the 24\textsuperscript{th} of August.\footnote{https://portal.msrc.microsoft.com/en-US/security-guidance/advisory/CVE-2018-8440}

Thus, by combining the timeliness of the Twitter information stream with the ability of deep neural architectures to accurately detect relevant tweets and identify useful pieces of information therein, OSINT-based threat intelligence platforms can improve significantly from the current state-of-art, to provide targeted, timely, and relevant threat intelligence.

\section{Conclusions}
This paper proposes deep neural network architectures to implement the core tasks of a processing pipeline to obtain timely, relevant and targeted security-related information from Twitter.
The proposed system is capable of gathering tweets from a set of accounts, filtering them based on a set of keywords defining an infrastructure to monitor, selecting the tweets containing relevant information, and identifying useful pieces of information in these tweets.
For that, we implemented convolutional and bidirectional long short-term memory neural networks.
We compare the performance of the proposed approach to well-established methodologies, verifying that the deep neural network architectures outperform those methodologies.
Three case studies specified by one nation-wide and two world-wide private organizations were used to validate the approach. 
Across the three case studies, the convolutional neural network binary classifier achieved an average TPR and TNR of 92\%, while the named entity recognition BiLSTM model achieved an average F1-score of 92\% in detecting specified labels.

Future research will focus on exploring multi-task learning architectures to shape our pipeline into a fully end-to-end neural network and to evaluate its impact on the models' performance and on the requirements for pipeline adaptation over time.

\section*{Acknowlegment}
This work was partially supported by the EC through funding of the H2020 DiSIEM project (H2020-700692), and by the LASIGE Research Unit (UID/CEC/00408/2019).

\bibliographystyle{IEEEtran}
\bibliography{bibliography}

\end{document}